\documentclass[letterpaper, 10 pt, journal, twoside]{IEEEtran}
\IEEEoverridecommandlockouts
\usepackage{tikz}
\usepackage{textcomp}
\usepackage{hyperref}

\newcommand\copyrighttext{%
  \footnotesize \textcopyright \centering 2021 IEEE. Personal use of this material is permitted. Permission from IEEE must be obtained for all other uses, in any current or future media, including reprinting/republishing this material for advertising or promotional purposes, creating new collective works, for resale or redistribution to servers or lists, or reuse of any copyrighted component of this work in other works.}

\newcommand\copyrightnotice{%
\begin{tikzpicture}[remember picture,overlay]
\node[anchor=south,yshift=5pt] at (current page.south) {\parbox{\dimexpr\textwidth-\fboxsep-\fboxrule\relax}{\copyrighttext}};
\end{tikzpicture}%
}

\usepackage{lipsum}
\usepackage{cite}
\usepackage{amsmath,amssymb,amsfonts}
\usepackage{algorithmic}
\usepackage{graphicx}
\usepackage{siunitx}
\usepackage[utf8]{inputenc}
\usepackage{textcomp}

\usepackage{amsfonts}
\usepackage{commath}
\usepackage{float}
\usepackage{enumitem}
\usepackage{epsfig}
\usepackage{epstopdf}
\usepackage[fleqn]{nccmath}
\usepackage[linesnumbered, ruled, vlined]{algorithm2e}

\def\BibTeX{{\rm B\kern-.05em{\sc i\kern-.025em b}\kern-.08em
    T\kern-.1667em\lower.7ex\hbox{E}\kern-.125emX}}
\newcommand{\Figref}{Fig.~\ref}

\makeatletter
\newcommand*{\rom}[1]{\expandafter\@slowromancap\romannumeral #1@}
\makeatother

\SetCommentSty{mycommfont}
\newcolumntype{P}[1]{>{\centering\arraybackslash}p{#1}}
\newcolumntype{M}[1]{>{\centering\arraybackslash}m{#1}}

\IEEEoverridecommandlockouts

\begin{document}

\markboth{IEEE Robotics and Automation Letters. Preprint Version. Accepted July, 2021}
{Schperberg \MakeLowercase{\textit{et al.}}: SABER: Data-Driven Motion Planner for Autonomously Navigating Heterogeneous Robots}

\raggedbottom
\title{SABER: Data-Driven Motion Planner for Autonomously Navigating Heterogeneous Robots}

\author{Alexander Schperberg$^{1}$, Stephanie Tsuei$^{2}$, Stefano Soatto$^{2}$, and Dennis Hong$^{1}$
\thanks{Manuscript received: March, 25, 2021; Revised June, 20, 2021; Accepted July, 22, 2021.}
\thanks{This paper was recommended for publication by Editor Hanna Kurniawati upon evaluation of the Associate Editor and Reviewers' comments. This work was supported by a grant (N00014-15-1-2064)  from  the  ONR}
\thanks{$^{1}$A. Schperberg, and D. Hong are with the Robotics and Mechanisms Laboratory, Department of Mechanical and Aerospace Engineering, University of
California, Los Angeles, CA 90095, USA.{\tt\small \{aschperberg28, dennishong\}@ucla.edu}}
\thanks{$^{2}$S. Tsuei, and S. Soatto are with the UCLA Vision Lab, Department of Computer Science, University of
California, Los Angeles, CA 90095, USA. {\tt\small \{stephanietsuei, soatto\}@ucla.edu}}

}

\maketitle
\copyrightnotice

\global\csname @topnum\endcsname 0
\global\csname @botnum\endcsname 0
\begin{abstract}
We present an end-to-end online motion planning framework that uses a data-driven approach to navigate a heterogeneous robot team towards a global goal while avoiding obstacles in uncertain environments. First, we use stochastic model predictive control (SMPC) to calculate control inputs that satisfy robot dynamics, and consider uncertainty during obstacle avoidance with chance constraints. Second, recurrent neural networks are used to provide a quick estimate of future state uncertainty considered in the SMPC finite-time horizon solution, which are trained on uncertainty outputs of various simultaneous localization and mapping algorithms. When two or more robots are in communication range, these uncertainties are then updated using a distributed Kalman filtering approach. Lastly, a Deep Q-learning agent is employed to serve as a high-level path planner, providing the SMPC with target positions that move the robots towards a desired global goal. Our complete methods are demonstrated on a ground and aerial robot simultaneously (code available at: \url{https://github.com/AlexS28/SABER}).

\end{abstract}



\section{Introduction}
\IEEEPARstart{T}{he} field of robotics has made remarkable progress in providing diverse sets of robotic platforms with different physical properties, sensor configurations, and locomotion capabilities (e.g., climbing, running, or flying). Thus, developing new planning algorithms that can be ubiquitously applied to a team of heterogeneous robots is a worthwhile endeavor, and applicable to a wide range of tasks from search and rescue to space exploration. However, for multi-agent planners to be used in unknown and uncertain environments, they should consider complex robot dynamics, uncertainty from imperfect exteroceptive and proprioceptive sensor measurements, update uncertainty when robots are in communication range, avoid obstacle collisions, and address desired multi-agent behavior.  

One common approach is to fully address some but not all of the described requirements while assuming the rest can either be satisfied in future work, or can be combined with other existing methods. However, combining multiple approaches towards a unified motion planning framework satisfying all requirements can be nontrivial, requiring close examination of the overall feasibility and performance of such a complex system. Thus, in this work, we will examine the feasibility and performance of an end-to-end motion planning framework that addresses the above requirements termed as $\lq$Stochastic model predictive control for Autonomous Bots in uncertain Environments using Reinforcement learning' or SABER.

\begin{figure}[!t]
    \centering
    \includegraphics[width=1\columnwidth]{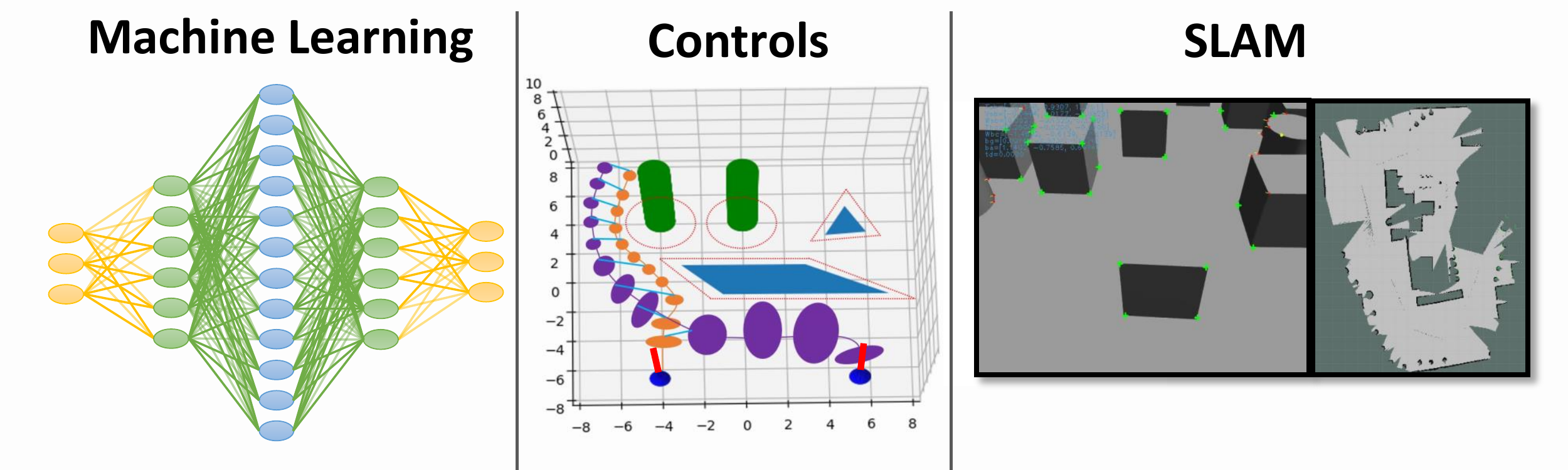}
    \caption{\textbf{SABER framework.} SABER combines controls (stochastic model predictive control), vision (simultaneous localization and mapping), and machine learning (RNN and DQN), to provide local and globally optimized solutions in unknown and uncertain environments.}
    \label{intro_pic}
\end{figure}

\subsection*{Summary of Our Contributions} 

(1) SABER is an end-to-end motion planning framework for a team of heterogeneous robots that unifies controls, vision, and machine learning approaches to plan paths that account for safety, optimality, and global solutions (our complete framework is shown on a UGV-UAV team).

(2) Cooperative localization algorithms are used for cross-communicating robots, which may include both non-Gaussian and Gaussian measurement noise, where uncertainty is modeled with recurrent neural networks (RNNs) for each agent's sensor configuration using outputs from simultaneous localization and mapping algorithms (SLAM).

(3) Instead of simple heuristics when sampling the map for target positions, we employ Deep Q-learning (DQN) for high-level path planning, which is easily modifiable for learning desired multi-agent behavior and finds global solutions (DQN scalability for more than two robots is also evaluated).

\begin{figure*}[!t]
    \centering
    \includegraphics[width=6.5in]{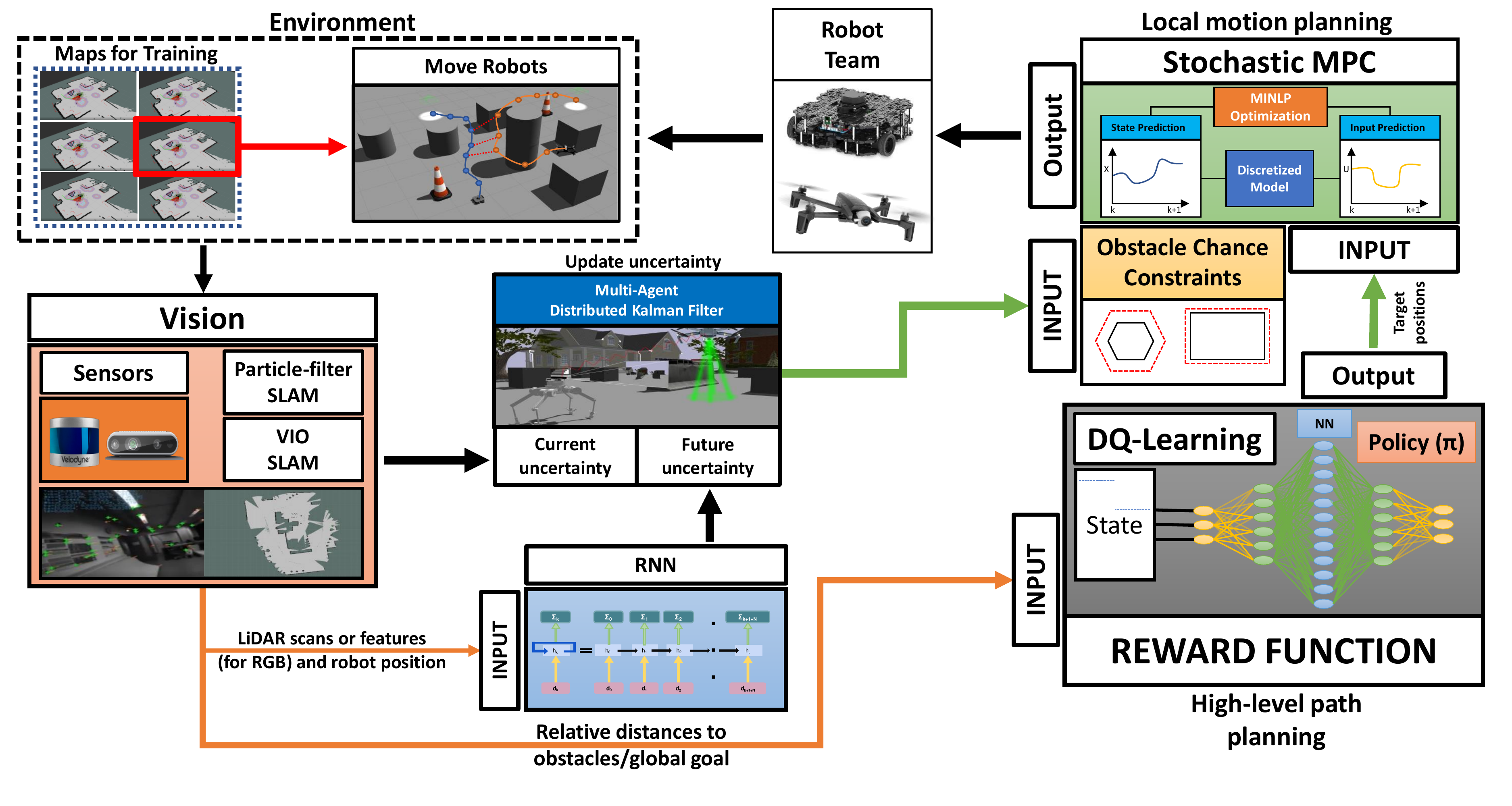}
    \caption{\textbf{SABER Algorithm.} This figure demonstrates the overall SABER planning algorithm in the testing phase, which can plan paths for one or more robots simultaneously. At timestep k, the environment provides information to robots that either carry a LiDAR or RGB camera and IMU; for the LiDAR configuration, a particle-filter SLAM is implemented, while for the RGB configuration, Visual-Inertial Odometry SLAM (VIO SLAM) is implemented. The sensors provide either scans or distance to feature information to a recurrent neural network model (which serve as inputs), and outputs the propagation of state uncertainty for future timesteps. If two or more robots are within communication range, a distributed Kalman filter updates the current and future states and their uncertainties to a more accurate estimate. These updated states and uncertainties are used to update the chance constraints for obstacle avoidance. These constraints are then considered by a stochastic MPC controller, which follows a given target position, provided by a deep Q-learning (DQN) agent that aims to move the robot towards a global goal. DQN uses the relative distances between the robots and the respective obstacles as its states, provides a target position for all robots as its actions, and is trained on several different maps with obstacles randomly distributed in each. Note, that the SMPC, SLAM, and RNNs components run on each robot individually, however, the DQN is run on a centralized base (which may be on the robot itself).} 
    \label{fig_scram}
\end{figure*}

\section{Related Works}
\label{relatedWorks}

Several works that examine planning for heterogeneous robots (typically composed of a two robot UGV-UAV team) have focused mainly on fusing different sensor data to build a unified global map \cite{heter_3}, integrating several components such as path planning, sensor fusion, mapping, and motion control towards a single framework \cite{heter_1}, or strictly analyzing multi-agent localization (i.e., multi-SLAM) \cite{heter_2}. While we also consider a UGV-UAV team as done in the above works, here, we are more concerned with the feasibility (i.e., computation time) of such a complex system, and also in how uncertainty is not only estimated for robots with different sensor configurations, but how it's tightly coupled with a local stochastic model predictive controller (SMPC) towards coordinating multi-agent behavior. 

We also seek to address the major challenge for multi-agent (or even single-agent) planners, which is to estimate a path that is both safe (e.g., considers uncertainty in agent/obstacle avoidance) and fast (e.g., finding the shortest path to the goal). This is significant because conservative approaches to safety would lead to over avoidance or non-optimal solutions, and high-risk behavior may cause undesired collisions. Currently, few motion planners fully investigate this problem. For example, in \cite{wurm_coordinating_2013}, the cost of reaching a target position for each robot in a heterogeneous team depends on its individual characteristics (e.g., varying sensors, travelling speed, and payloads). However, by not considering uncertainty in their planner, their cost-to-go function can be significantly affected by disturbances. Conversely, a multi-agent planner that does consider probabilistically-safe motion planning can be found in \cite{bajcsy_scalable_2019}. Still, their planner may lead to conservative solutions as they assume a worst-case behavior approach to safety. SABER addresses the problem of avoiding obstacles without over avoidance by using an RNN, which predicts and propagates future state uncertainty dynamically and does not make the assumption that uncertainty increases when no future measurements are received \cite{Schperberg_rnn}. The downside with an RNN (which is typical with learning-based approaches) is that the accuracy of the uncertainty estimates is directly correlated with the quality of the data collection. 

The geometric representation of obstacles is also critical for planning. For example, FASTER \cite{tordesillas_faster_2019} is a decentralized and asynchronous planner where obstacles are represented as outer polyhedrals (estimated from convex decomposition) and applied as constraints into the optimization. In SABER, we also represent obstacles as polyhedral constraints for each timestep, however, we decompose them into a disjunction of linear chance constraints (thus, obstacle $\lq$size and location' are a function of exteroceptive and proprioceptive uncertainty). Using chance constraints in motion planning is not new, and has been shown with success in single-agent planners such as \cite{blackmore_chance-constrained_2011} and \cite{risk_aware}. In this work, our chance constraints are also influenced by the cross-communication uncertainty of a heterogeneous robot team (see \ref{methods:coop}). Additionally, while obstacle constraints can be explicitly used in the optimization, other works show a learning-based approach to avoid collisions by modeling the distribution of promising regions for travel \cite{region_learning} or predicting the separation distance between the robot and its surroundings \cite{chase_kew_neural_2021}. Our work is a hybrid approach, where we use the RNN to predict uncertainty of state estimates (which affect the $\lq$size' of polyhedral obstacles), but still use these obstacles as constraints in our SMPC optimization. This choice sacrifices computation time, but may be more generalizable to environments not observed in training and prevent collisions.

Finding a suitable path to the goal also has a wide array of different solutions. Most commonly, sampling-based methods which do not consider workspace topology (such as grid-sampling \cite{grid_sample} or rapidly-exploring random tree or RRT and its variants \cite{gammell_informed_2014}) can lead to very dense roadmaps and may not scale well when the shortest path to the goal is desired. This issue has been investigated in \cite{probabilistic_roadmaps}, which uses a self-supervised learning approach to build sparse probabilistic roadmaps (PRM) for bias sampling (sampling only regions the robot is likely to safely travel). Moreover, using a learning-based approach for path planning also has the potential for integrating semantic behavior that can be gained from multi-agent coordination, as evidenced in \cite{multi_agent_rendezvous} and \cite{Baker2020Emergent}. Motivated by these works, we use a DQN for high-level planning, where simple modifications of a reward function can yield desired multi-agent behavior (e.g., rewarding agents based on proximity or reaching the goal concurrently). Nevertheless, the tradeoff of using a DQN compared to sampling-based methods is that it cannot guarantee asymptotic optimality (e.g., RRT-star) or probabilistic completeness (e.g., PRM). However, for our DQN, we are primarily concerned with the feasibility relative to our application (i.e., finding a near-optimal path in a computationally efficient manner while satisfying multi-agent behavior).

\begin{algorithm}[!ht]
	\SetAlgoLined
	\DontPrintSemicolon
	Initialize state $X_{k}^{i:n_{r}}$, goal $X_{goal}^{i:n_{r}}$, $dt$, horizon $N$, robot size $r_{i:n_{r}}$, timestep $k$, empty $Map^{i:n_{r}}$, uncertainty $\Sigma_{k:k+N+1}^{i:n_{r}}$, error to goal $\epsilon$, number of robots $n_r$ \\
	\BlankLine
	\While{$ \| X_{k}^{i:n_{r}} - X_{goal}^{i:n_{r}} \|_2 > \epsilon$}{
	    
	    \textbf{if} LiDAR configuration: \\
        \ \ $X_{k}^{i:n_{r}}, \Sigma_{k}^{i:n_{r}}, Map^{i:n_{r}} \leftarrow \text{Particle-filter SLAM}(\text{odom, scans, } Map^{i:n_{r}})$ \\
        \ \ $\Sigma_{k+1:k+N+1}^{i:n_{r}} \leftarrow \text{RNN}^{i:n_{r}}(X_{k}^{i:n_{r}}, \text{scans})$ \\
        
        \textbf{if} RGB camera configuration:
        \ \ $X_{k}^{i:n_{r}}, \Sigma_{k}^{i:n_{r}}, Map^{i:n_{r}} \leftarrow \text{VIO SLAM}(\text{IMU, RGB, } Map^{i:n_{r}})$ \\
        \ \ $\Sigma_{k+1:k+N+1}^{i:n_{r}} \leftarrow \text{RNN}^{i:n_{r}}(X_{k}^{i:n_{r}},\ \text{features})$ \\
        
        \textbf{continue:}\\
        $X_{ref}^{i:n_{r}} \leftarrow \text{Deep Q-Learning}(X_{k}^{i:n_{r}}, X_{goal}^{i:n_{r}}, Map^{i:n_{r}})$ \\
        
        
        $\mathcal{O}_{j:n_{obs}} \leftarrow \text{checkObstacles}(X_{k}^{i:n_{r}}, r_{i:n_{r}}, Map^{i:n_{r}})$ \\
        
        $X_{k+1:k+N+1}^{i:n_{r}}, U_{k+1}^{i:n_{r}} \leftarrow \text{SMPC}(X_{ref}^{i:n_{r}}, X_{k}^{i:n_{r}}, \mathcal{O}_{j:n_{obs}}, \Sigma_{k:k+N+1}^{i:n_{r}})$ \\
        \textbf{if }$\forall robot \ i,j:n_{r} \text{ within communication range:}$ \\
       \ \ $\Sigma_{k:k+N+1}^{i:n_r},X_{k:k+N+1} \leftarrow \text{CoopLocalization}(\Sigma_{k:k+N+1}^{i:n_{r}}, X_{k:k+N+1}) $\\
	}
	\caption{SABER}
	\label{SABER}
\end{algorithm}

\section{Methods}

The SABER framework contains both learning (requiring data collection) and non-learning components (traditional control schemes). The non-learning components consist of an SMPC and a distributed Kalman filter, while the learning components consist of an RNN and DQN agent. The RNN and DQN components are trained separately and offline before being implemented into the overall system for online deployment (note, that the RNN is supervised by the SMPC controller on each robot, while the DQN is not integrated with any other component during training). Overall, the algorithm is structured as an SMPC problem, which moves a robot toward a target location as formulated in \ref{methods:smpc}. By using state uncertainties and obstacle locations, obstacles are represented as chance constraints within the SMPC cost function (\ref{methods:smpc:cc}, \ref{methods:smpc:cc2}). If two or more robots are in communication range, their state and uncertainty values are updated using a distributed Kalman filtering approach as described in \ref{methods:coop}. To quickly propagate state uncertainties for future timesteps, we use different RNN models based on the robot's sensor configuration, as explained in \ref{methods:rnn}. In \ref{methods:dqn}, we formulate a DQN approach, providing the SMPC with target locations which help generate trajectories that move the robots toward a global goal and prevent local minima solutions. See Algorithm \eqref{SABER}, \Figref{fig_scram}, or the attached video\footnote{\url{https://youtu.be/EKCCQtN5Z6A}} for an overview of the methods, and \ref{implementation} for implementation details.

\subsection{Stochastic Model Predictive Control Formulation}
\label{methods:smpc}
The goal of the cost function (equation \eqref{objfunction}) is to find the optimal control value $U_{k}$ that minimizes the distance between the current and predicted states ($X_{k\rightarrow N+1}$) with a reference state or trajectory ($X^{(ref)}$) while under equality and inequality constraints -- where $X_{k}^{i}$ is given by the results of localization for each robot $i$, and $k$ is the current timestep ($Q$ and $R$ are control matrices, and $P$ is described further below): 

\mathchardef\mhyphen="2D

\begin{equation}
\label{objfunction}
    \begin{split}
        \min_{U_{k:k+N\mhyphen1}^{i}} \sum_{k}^{k+N-1} \lVert X_{k+1}^{i}-X_{k+1}^{(ref)i} \rVert^{2}_{Q^{i}}
        + U_{k}^{i\top}R^{i}U_{k}^{i} \\ + \lVert X_{k+N+1}^{i}-X_{k+N+1}^{(ref)i} \rVert^{2}_{P^{i}}
    \end{split}
\end{equation}

\begin{alignat}{2}
& \text{s.t.} \quad
& & {X_{k+1}^{i}=f^{i}(X_{k}, U_{k})=A^{i} X_{k}^{i}+B^{i}U_{k}^{i}}+W_k^{i} \\
&&& X_{k}^{i} \sim \mathcal{N}(\Bar{X_{k}^{i}}, \Sigma_k^{i}), W_k^{i} \sim \mathcal{N}(0, \sigma^{2i}) \\
&&& X_{limit}^{i} \geq |X_{k}^{i}|, U_{limit}^{i} \geq |U_{k}^{i}| 
\end{alignat} 
\textit{Obstacle Constraints:}
\begin{gather}
\text{Pr}\left(\bigwedge_{j=1}^{N_{O}} \mathcal{O}_{j}\right) \geq 1-\Delta 
\end{gather}

Constraint (2) represents multiple shooting constraints, which ensure that the next state is equivalent to a time-invariant linear discretized model, where $A$ and $B$ represent the robot's dynamic matrices. Uncertainty in state as well as the addition of a non-unit variance random Gaussian noise ($W$) is shown in (3). Note, that the propagation in state uncertainty ($\Sigma_{k+1\rightarrow N+1}^{i}$) is received from an RNN model (Section \ref{methods:rnn}), and can be affected by cooperative localization algorithms (Section \ref{methods:coop}) at timestep $k$, if multiple robots are in communication range at timestep $k$.

Limits on state and controls variables are imposed by constraint (4). For robustness \cite{rawlings_model_2017}, a terminal cost is included with a weighting matrix $P$ which can be obtained by solving the discrete-time Riccati equation \eqref{ricatti}:
\begin{equation}
    \label{ricatti}
    \begin{split}
A^{i\top}P^{i}A^{i} - P^{i} - A^{i\top}P^{i}B^{i}(B^{i\top}P^{i}B^{i}+R^{i})^{-1}B^{i\top}P^{i}A^{i}+\\Q^{i}=0 
    \end{split}
\end{equation}
\subsubsection{Chance Constraints for Obstacle Avoidance}
\label{methods:smpc:cc}
Constraint (5) represents chance constraints that enable obstacle avoidance subject to uncertainty in convex regions as done in \cite{blackmore_chance-constrained_2011}. This constraint can be rewritten as a disjunction of linear constraints for obstacle $\mathcal{O}_{j}$:

\begin{equation}\mathcal{O}_{j} \Longleftrightarrow \bigvee_{k \in T\left(\mathcal{O}_{j}\right)} \bigwedge_{i \in G\left(\mathcal{O}_{j}\right)} a_{i}^{{\top}} \Bar{X}_{k} - b_{i} \geq c_{i}\end{equation}
where  $G\left(\mathcal{O}_{j}\right)$ is the set of linear constraints (indexed by $i$) for each obstacle (indexed by $j$), $T\left(\mathcal{O}_{j}\right)$ is the set of timesteps in the MPC prediction horizon (indexed by $k$), $a_{i}$ is the vector normal to each line constraint and directed toward state $\Bar{X}_{k}$, $r$ is the radius/size of the robot, and $c_{i}$ is given by:

\begin{equation}c_{i}=\sqrt{2 a_{i}^{\top} \Sigma_{k} a_{i}} \cdot \operatorname{erf}^{-1}(1-2 \delta_{j}), \ \delta_{j} \leq 0.5 \end{equation}

Important to consider is that the degree of $\lq$risk' can be controlled by changing the values of $\delta_{j}$ (related to $\Delta)$ for each obstacle $\mathcal{O}_{j}$. Lower values lead to more evasive behavior (robot moves further away from obstacles) while higher values lead to more risky behavior (robot moves closer to obstacles). 

If obstacles are assumed circular (centered at $x_{o_j}$, $y_{o_j}$), we can use the following equation, where $a_{i}$ is equal to an identity vector, $x_{k}$ and $y_{k}$ is the center position of the robot, and only a single $c_{j}$ value needs to be calculated per obstacle:

\begin{equation}-\sqrt{\left(x_{k}-x_{o_j}\right)^{2}+\left(y_{k}-y_{o_j}\right)^{2}}+(r+ c_{j}) \leq 0 \end{equation}

\subsubsection{Mixed-Integer Nonlinear Programming}
\label{methods:smpc:cc2}
To more effectively consider the disjunctive convex program for polygon-shaped obstacles, as introduced by (7), we can change these constraints into a mixed integer format (we assume the line constraints are in the x-y plane, however, the same equations can be used for the x-z, and y-z planes respectively):

\begin{equation}\mathcal{O}_{j} \Longleftrightarrow \bigvee_{k \in T\left(\mathcal{O}_{j}\right)} \bigwedge_{i \in G\left(\mathcal{O}_{j}\right)} I_{i,j} \text{dist}(\Bar{X}_{k}, a_{i}, m_{i}, b_{i}) \geq I_{i,j}( r+c_{i})\end{equation}

\begin{equation} 
x_{l} = a^{*}_{i}x_{k}-y_{k}+b_{i} / (a^{*}_{i} - m_{i})
\end{equation}
\begin{equation}
y_{l} = m_{i}x_{k}+b_{i}
\end{equation}

\begin{equation}\text{dist(*)}=\left|-m x_{k}+y_{k}-b_{i}\right| / \sqrt{m_{i}^{2}+1}\end{equation}

\begin{equation}\text {dist(*)}\left\{\begin{array}{ll}{\textbf{IF } \text{sign}(y_{l} - y_{k}) = \text{sign}(a_{y})} \bigvee \\ \ \  \ \  \text{sign}(x_{l}-x_{k})=\text{sign}(a_{x}), & {-\text{dist(*)}} \\ {\textbf{ELSE}} & {\text{dist(*)}}\end{array}\right.\end{equation}

\begin{equation}I_{i,j}=\{0,1\} \forall i,j\end{equation}

\begin{equation}\sum_{i=1}^{size(I_{j})} I_{i,j} \geq 1 \quad \forall j\end{equation}
where $m_{i}$ and $b_{i}$ are the slope and y-intercept of each line constraint ${i}$ belonging to obstacle $\mathcal{O}_{j}$, $a^{*}_{i}$ is $a_{y}/a_{x}$, $x_{k}$ and $y_{k}$ are the $x$ and $y$ position retrieved from robot state $\Bar{X}_{k}$, and the coordinates of the point on the line constraint closest to $\Bar{X}_{k}$ is represented by $x_{l}$ and $y_{l}$ (equations (11) and (12)). The dist(*) function approximates the distance between $\Bar{X}_{k}$ and one of the linear constraints of an obstacle as shown in equation (13). Equation (14) returns a positive distance if the robot is $\lq$outside' the obstacle boundary, and a negative distance if the robot is $\lq$inside' the obstacle boundary (a negative distance would cause the line constraint to fail). By definition of constraint (10), only one or more of the line constraints need to be satisfied per obstacle $\mathcal{O}_{j}$, which is ensured by using binary integer variables under constraints (15) and (16) (e.g., for line constraint $i$ belonging to $\mathcal{O}_{j}$, if $I_{i,j}=1$, the robot is outside this obstacle). For simplicity, we assume we have a $\lq$perfect' object detection system. If the robot is close enough to an obstacle, the obstacle is automatically $\lq$seen' and embedded into the SMPC cost function.

\subsection{Cooperative Multi-Agent Localization}
\label{methods:coop}
While the propagation of uncertainty for each robot is calculated using an RNN (see \ref{methods:rnn}), updating the uncertainty after information is exchanged with another robot is done using a distributed Kalman filtering approach \cite{coop_local2000}. Thus, when two or more robots are in communication range (as pre-specified by the user), their individual uncertainty estimates should be updated to correctly reflect the gain from additional sensor information. 

Equations (17) - (25) describe how the pose for robot $i$ is updated while in communication range of another robot $j$. The same equations can be further extrapolated to consider additional robots as explained in \cite{coop_local2000}.\\
\textit{For $\forall i,j$ and $k \rightarrow k+N+1$:}  \\
\textit{Propagation}: 
\begin{equation}
\Sigma_{k+1}^{i}, \Sigma_{k+1}^{j} = RNN^{i}(*), RNN^{j}(*)
\end{equation}
\begin{equation}
X_{k+1}^{i}, X_{k+1}^{j} = f^{i}(X_{k}^{i}, U_{k}^{i}), f^{j}(X_{k}^{j}, U_{k}^{j})
\end{equation}
\textit{Update}:
\begin{equation}
\Bar{X}_{k+1}^{+i} = \Bar{X}_{k+1}^{i} + K_{k+1}^{i}(Z_{k+1}^{ij}-(\Bar{X}_{k+1}^{i}-\Bar{X}_{k+1}^{j}))  
\end{equation}
\begin{equation}
S_{k+1}^{ij} = \Sigma_{k+1}^{i} + \Sigma_{k+1}^{j} + R_{k+1}^{ij}
\end{equation}
\begin{equation}
Z_{k+1}^{ij} = X_{k+1}^{i} - X_{k+1}^{j} 
\end{equation}
\textit{Update A (first time robots meet)}:
\begin{equation}
\Sigma_{k+1}^{+ij} = \Sigma_{k+1}^{i}(S_{k+1}^{ij})^{-1}\Sigma_{k+1}^{j}
\end{equation}
\begin{equation}
K_{k+1}^{i} = \Sigma_{k+1}^{i}(S_{k+1}^{ij})^{-1}
\end{equation}
\textit{Update B (all other times robots meet)}:
\begin{equation}
\Sigma_{k+1}^{+ij} = \Sigma_{k+1}^{ij}-[\Sigma_{k+1}^{i}-\Sigma_{k+1}^{ij}](S_{k+1}^{ij})^{-1}\Sigma_{k+1}^{j}[\Sigma_{k+1}^{ij}-\Sigma_{k+1}^{j}]
\end{equation}
\begin{equation}
K_{k+1}^{i} = (\Sigma_{k+1}^{i}-\Sigma_{k+1}^{ij})(S_{k+1}^{ij})^{-1}
\end{equation}
where $Z_{k+1}^{ij}$ is the relative pose measurement between robot $i$ and robot $j$ ($X_{k}$ is received by current localization, and $X_{k+1\rightarrow k+N+1}$ can be retrieved by the SMPC solution), and $R^{ij}$ is the relative measurement noise between robot $i$ and robot $j$.

\subsection{Recurrent Neural Network for Uncertainty Propagation}
\label{methods:rnn}
\begin{figure}[!t]
    \centering
    \includegraphics[width=1.0\columnwidth]{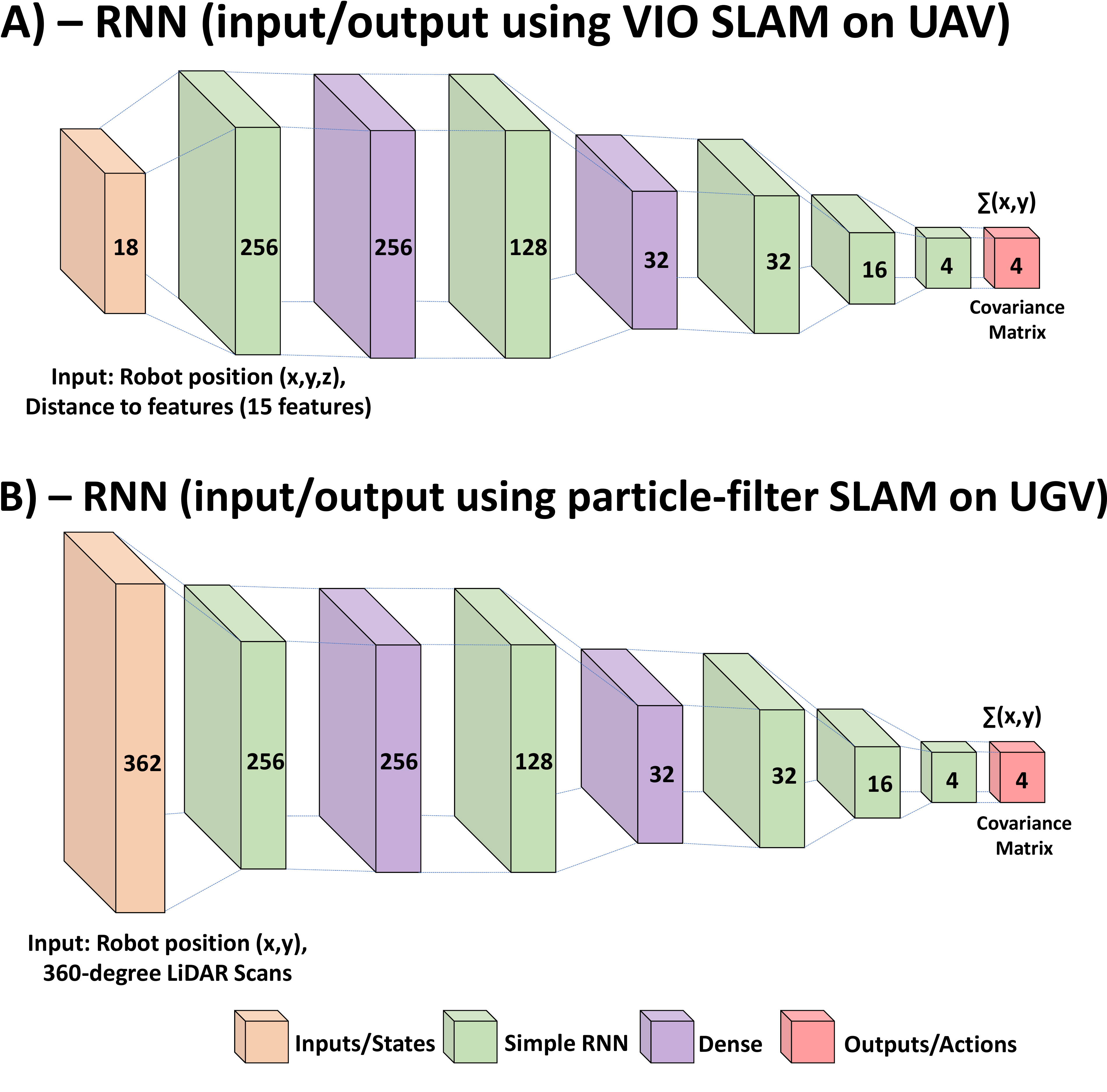}
    \caption{\textbf{Network structures.} We show the RNN structure used to model an EKF from a VIO SLAM algorithm in (\textit{A}), or a particle-filter SLAM algorithm in (\textit{B}) (\ref{methods:rnn}). The inputs are shown in orange, and correspond to either features/robot position (using VIO SLAM) or LiDAR scans/robot position (using particle-filter SLAM). The outputs are shown in red, and correspond to the x-y covariance matrix (which represents uncertainty in x-y position). The layer type is color coded below, where green represents a simple RNN layer, and purple a dense layer.}
    \label{networks}
\end{figure}
\begin{figure}[!t]
    \centering
    \includegraphics[width=1.0\columnwidth]{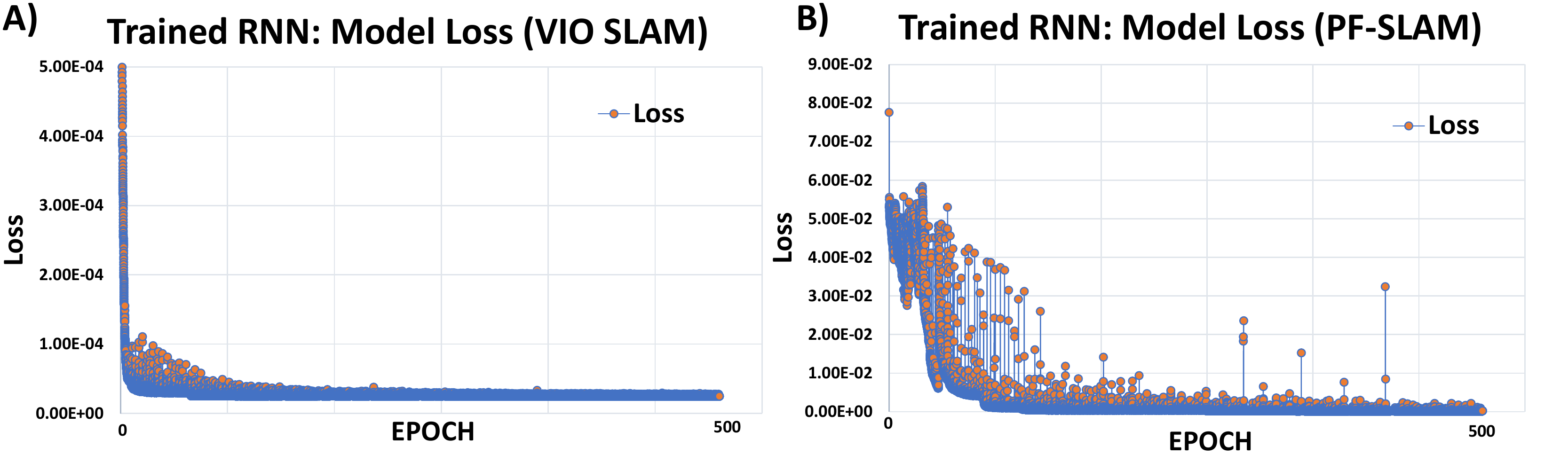}
    \caption{\textbf{Training loss.} Here we show the training loss for the RNNs, which were trained on uncertainty covariance outputs (in position) of a Visual-Inertial SLAM in (\textit{A}) and a particle-filter SLAM in (\textit{B}). The training was done using 500 epochs and on 4 different maps. Note, that the noise observed in (\textit{B}) may be due to the particle-filter estimations/simplifications done in \cite{Gmapping}.}
    \label{rnn_loss}
\end{figure} 

Because our SMPC calculates the optimal control ($U_{k}$) based on a prediction horizon ($X_{k+1\rightarrow k+N+1}$, $U_{k\rightarrow k+N}$), we must also provide as input the propagation of uncertainty ($\Sigma_{k+1 \rightarrow k+N+1}$) at each timestep. As described in more detail in \cite{Schperberg_rnn}, an RNN can provide a computationally fast prediction of future state uncertainties (as it only requires a small network) and can operate in continuous space, making it ideal for online replanning in complex environments. This is achieved as the RNN can model the behavior of a filter (e.g., particle filter or EKF) from SLAM algorithms. However, in this study, we have multiple robots with different sensor configurations, which requires training separate RNN models for each. 

In this work, we estimate state uncertainty using two different SLAM algorithms, a Rao-Blackwellized particle-filter SLAM (LiDAR camera configuration) and a Visual-Inertial Odometry (VIO) SLAM (RGB camera configuration). In the particle-filter case, the following equation is used for factorization:
\begin{equation}\begin{array}{c}
p\left(x_{1: k}, m \mid z_{1: k}, u_{1: k-1}\right)= \\
p\left(m \mid x_{1: k}, z_{1: k}\right) \cdot p\left(x_{1: k} \mid z_{1: k}, u_{1: k-1}\right)
\end{array}\end{equation}
where $x_{1:k}$ = $x_{1},...,x_{k}$ is the robot's trajectory, $z_{1:k}$ = $z_{1},...,z_{t}$ are the given observations, $d_{1:k-1}$ = $d_{1},...,d_{k-1}$ are the odometry measurements, and $p(x_{1: k}, m \mid z_{1: k}, d_{1: k-1})$ is the joint posterior estimate about map $m$ (see \cite{AMCL}). 
For training the RNN for the particle-filter SLAM case, we have 362 input units for each timestep $k$. The first 2 units is the center position of the robot ($x$ and $y$), and the 360 other units represent the range distances from LiDAR scans (e.g., for timestep k, we have $d_{k}^{1:360}$ relative scan distances). The output layers, which use a linear activation function, correspond to the robot's 2$\times$2 x-y covariance matrix which is flattened into a 4$\times$1 array or 4 output units. For the VIO SLAM configuration, we used the same methods for training the RNN as described in \cite{Schperberg_rnn}. See \Figref{networks} for an overview of the network structure, and \ref{implementation} for further implementation details.

\subsection{Deep Q-Learning (DQN) for Global Planning}
\begin{figure*}[!t]
    \centering
    \includegraphics[width=400px]{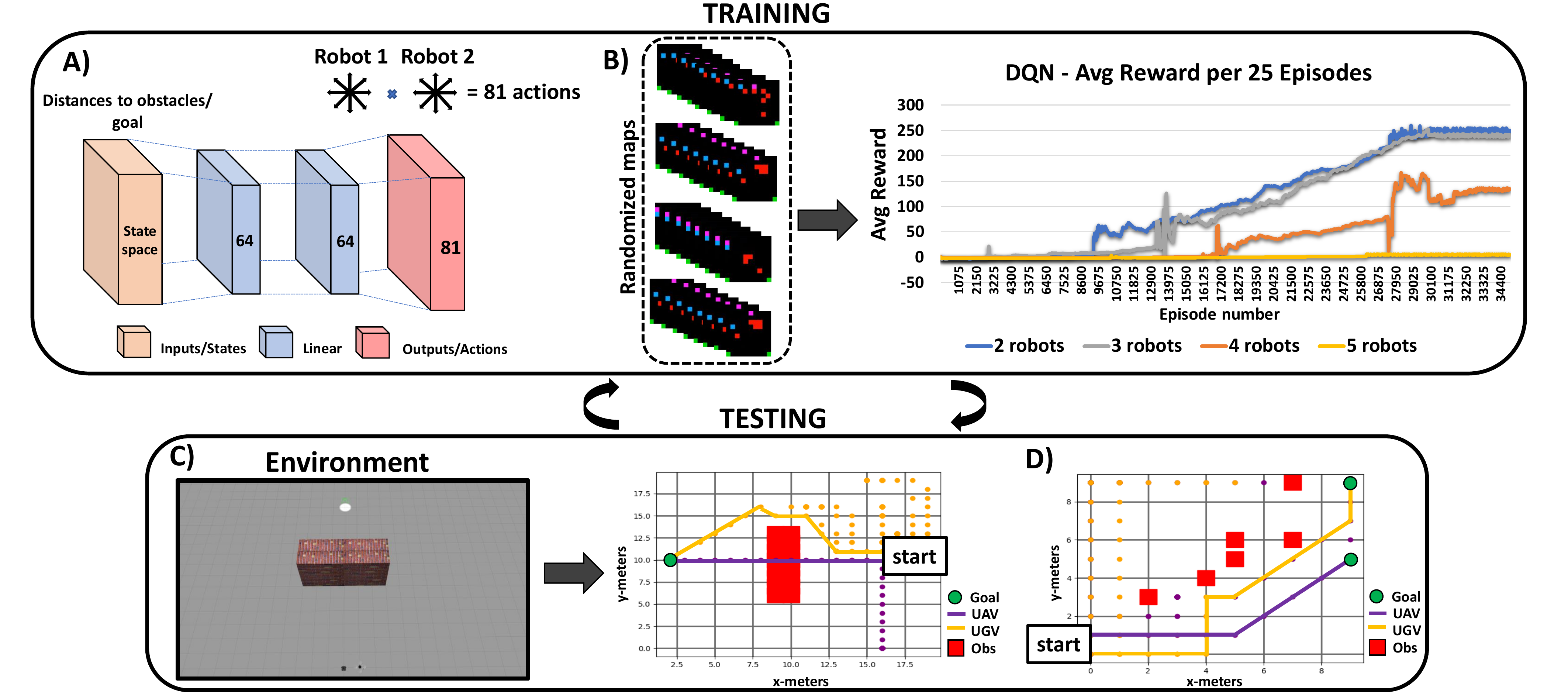}
    \caption{\textbf{DQN Training and Testing Procedure.} In (\textit{A}), we show the neural network structure used in our DQN algorithm (\ref{methods:dqn}). The network maps the inputs (i.e., states or relative distances between robots and obstacles/goals) to the outputs (i.e., actions or next target positions for the robots). The states and actions are connected by a linear neural network model (blue). In (\textit{B}) we visually show the training process of the DQN for a 2 to 5 robot team, where all robots were trained to go to the goal location while avoiding obstacles (obstacles are randomized for each episode). The average rewards (calculated from 25 episodes at a time and divided by number of robots) are shown across the 35,000 episodes of training (training time was 5 hours). In (\textit{C}) we show an example of how the environment can be transcribed into a 2D plot and apply the DQN to traverse multiple robots toward their goals. We also allow the UAV to fly over the obstacles (and assume we know the height of the obstacle \textit{a priori}) while the UGV must avoid it. In (\textit{D}) we show another example of our DQN, but this time the 2-robot UAV and UGV team have separate goal locations, and we add a reward incentive when both robots are near each other at each timestep.} 
    \label{fig_dqn_methods}
\end{figure*}
\label{methods:dqn}

\subsubsection{DQN formulation}

To provide local target positions ($X_{ref}^{i:n_{r}}$) that direct the robots toward a global goal ($X_{goal}^{i:n_r}$), we implement a DQN agent and use the Bellman equation:
\begin{equation}\label{bellman}Q(s_{k}, a_{k})=r+\gamma \max _{a_{k+1}} Q\left(s_{k+1}, a_{k+1}\right)\end{equation}
where the state is represented by $s_{k} \in \mathbb{R}^{({N}_{j}\times n_{r})+n_{r}}$, action by $a_{k} \in \mathbb{R}^{9^{n_{r}}}$, learning rate by $\gamma$, $n_{r}$ is the number of robots, and reward function by $r$ (where ${N}_{j}$ is the number grid spaces required to represent each obstacle; described further below). The idea of DQN is to use the Bellman equation \eqref{bellman} and a function approximator (with neural networks) to reduce the loss function (we use the Adam optimizer) \cite{Q-learning}.

Ultimately, the goal of the DQN-agent is to generate target positions at each timestep k for multiple robots and to move them towards a global goal position. To accomplish this and generalize our methods to any map (or changing the location of obstacles after each episode), we use the relative distances between the robot and surrounding obstacles and the relative distance between the robot's position and the global goal as our states: $s_{k} = (d_{j}^{i:n_{r}}, d_{j+1}^{i:n_{r}}, d_{j+2}^{i:n_{r}},...d_{N_{O}}^{i:n_{r}}, d_{goal}^{i:n_{r}})$, where $d_{j}^{i:n_{r}}$ is the relative distance between the robot and obstacles and can be described by $\lVert X_{k}^{i:n_{r}}-\mathcal{O}_{j}^{i:n_{r}}\lVert$, and $d_{goal}^{i:n_{r}}$ by $\lVert X_{k}^{i:n_{r}}-X_{goal}^{i:n_{r}}\lVert$ ($X_{k}$ is assumed to be the $x$ and $y$ position of the robot). Our DQN agent is trained on a 2D-grid map, where the robots and obstacles are represented as squares (1m$^{2}$) within the grid. Thus, the actions permitted by the robot is an 8-directional x-y movement (or no movement) at each timestep. The reward function is simply formulated by providing a positive reward $(+1)$ if the robot gets to the goal and a negative reward $(-1)$ if the robot hits an obstacle, which results in termination of the episode (we allow the UAV to $\lq$fly' over some obstacles by modifying its reward function to not receive a penalty for hitting that obstacle). To test different behaviors, we also added a reward $(+0.1)$ at each timestep when both robots are within a 2 meter distance (see (D) in \Figref{fig_dqn_methods}).

We also assume the dimensions of each obstacle are known \textit{a priori} (to estimate this without this assumption may require an object detection pipeline). Thus, by knowing the height of each obstacle, a UAV can use this value in its chance constraint to fly over obstacles. Finally, mapping our states to our actions is done through a linear neural network. For an overview of the network structure, and the training/testing process, see \Figref{fig_dqn_methods}.


\section{Experimental Validation}
\begin{figure}[!t]
    \centering
    \includegraphics[width=1.0\columnwidth]{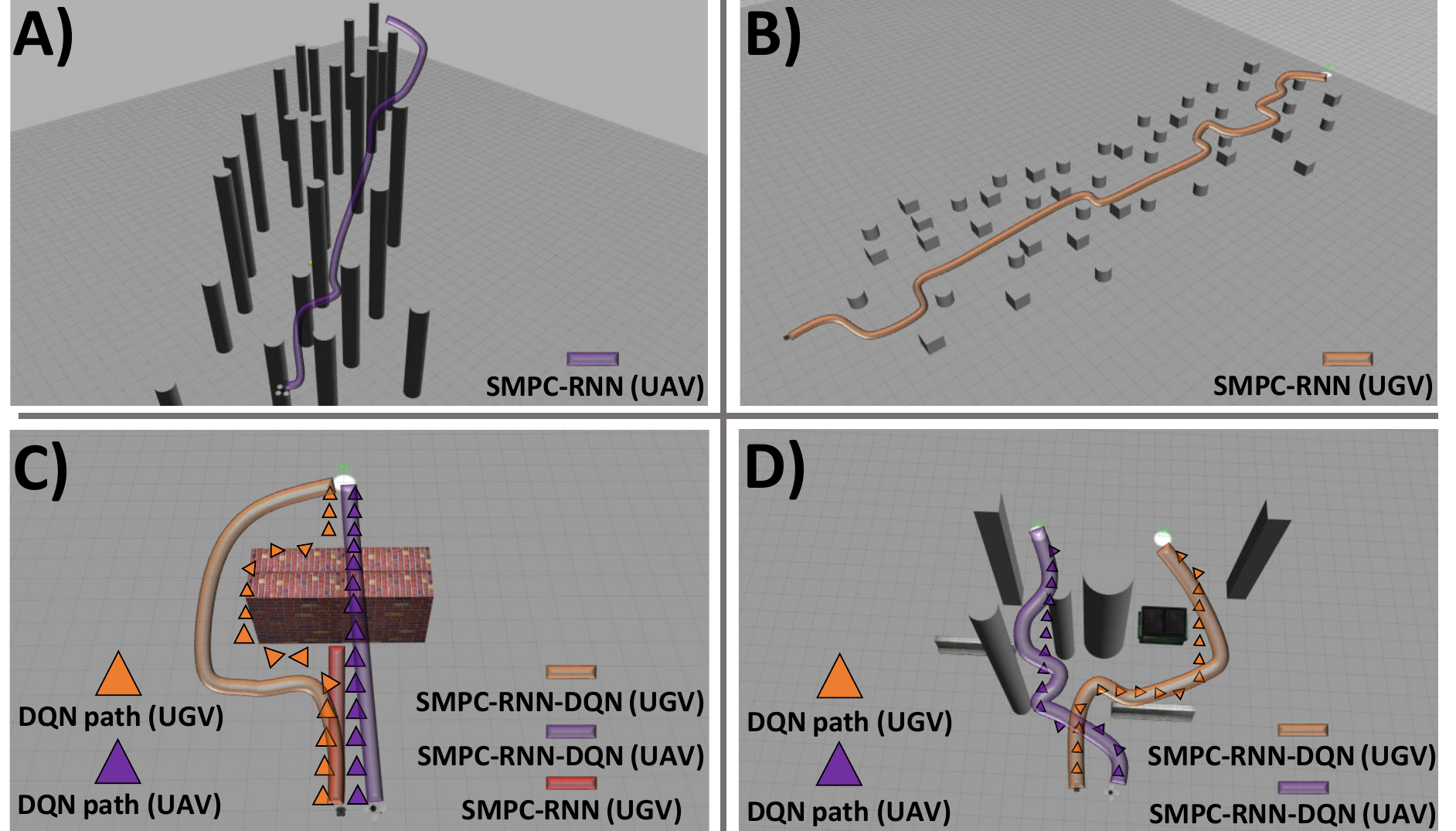}
    \caption{\textbf{SABER Algorithm Results.} This figure demonstrates the overall SABER planning algorithm. In (\textit{A}) and (\textit{B}) we first show the capability of the SMPC-RNN to navigate the UGV and UAV in a densely populated space. In (\textit{C}), we show that the SMPC-RNN of the UGV cannot get to the goal state, because of the occurrence of a local minima. However, with a DQN (which provides a global path illustrated by triangles), the UGV (orange) can correctly maneuver around the obstacle. The UAV (purple) can simply use it's z-axis to fly above the obstacle. A more complex example is shown in \textit{(D)}, where both robots are directed towards different goal locations simultaneously.}
    \label{fig_saber}
\end{figure}
\begin{figure}[!t]
    \centering
    \includegraphics[width=0.83\columnwidth]{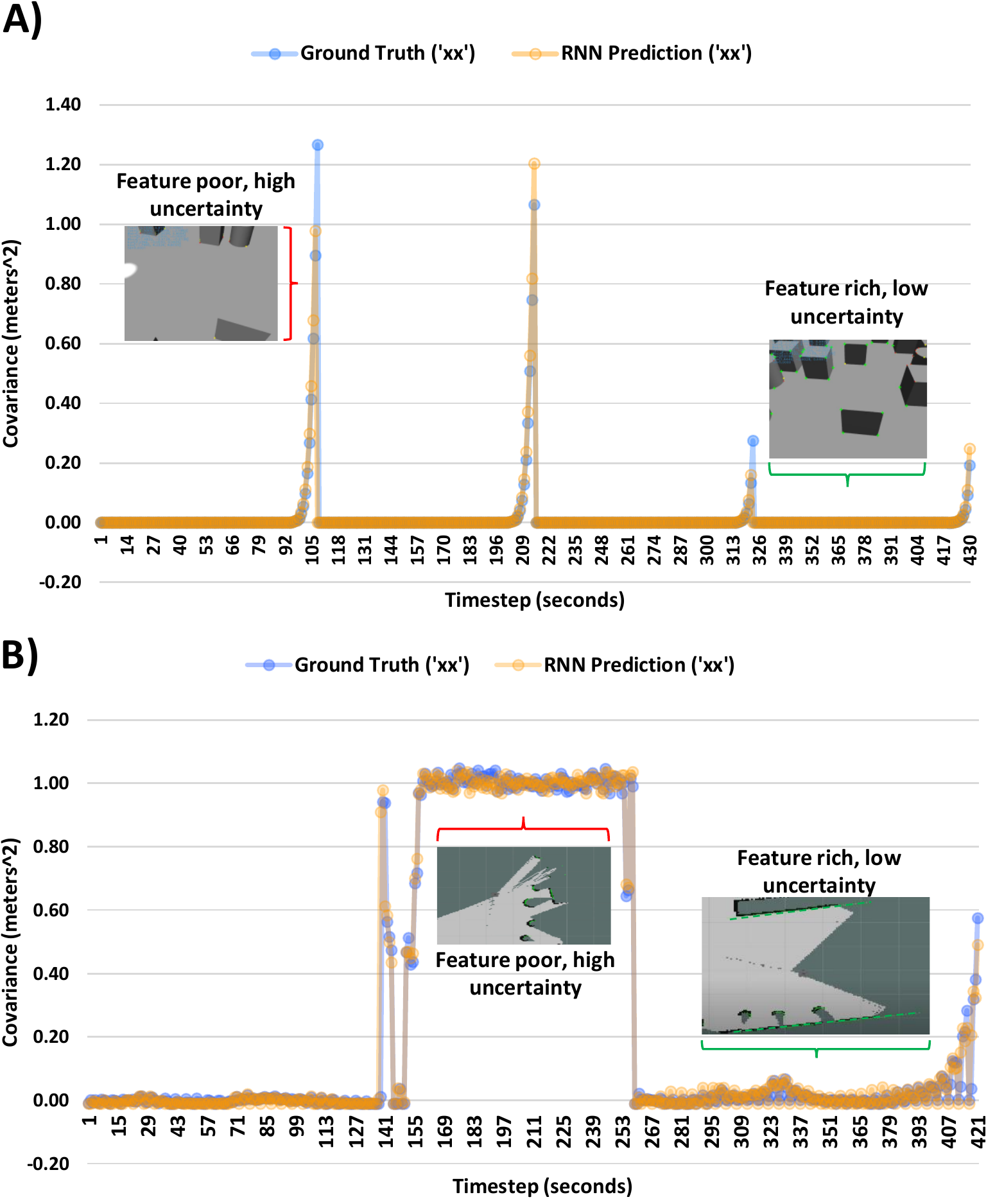}
    \caption{\textbf{RNN Results.} In this figure, we show the true covariance value (where $\lq$xx' represents the covariance of the center of mass in the x position as an example) in blue and the predicted covariance in orange for about 430 seconds of data. This is done when modeling the uncertainty using VIO SLAM (\textit{A}) and particle-filter SLAM (\textit{B}) on a test map. Note, that the predicted and true values almost perfectly align, demonstrating the RNN's ability to make valid uncertainty predictions. Lastly, we show more explicitly in the graphs, that when more features are tracked (green arrow) the lower the estimated uncertainty, while fewer features corresponded to higher uncertainty (red arrow).}
    \label{rnn_results}
\end{figure}

\subsection{Implementation details}
\label{implementation}
SABER is demonstrated on a UGV (Turtlebot3) equipped with a 360 degree LiDAR camera, and a UAV (Quadrotor) equipped with a RGB camera in a Gazebo simulation running in real-time with additive noise. The dynamic equation of motion for the UGV assumes states composed of center of mass position and heading angle ($X=[x, y, \theta]$), actions composed of linear and angular velocity ($U=[v, \omega]$) and the following matrices for the discretized equation:

\begin{equation*}A = \mathbb{I}\in \mathbb{R}^{3\times3}, \  B=\left[\begin{array}{cc}
\cos (\theta) \delta t & 0 \\
\sin (\theta) \delta t & 0 \\
0 & \delta t
\end{array}\right]\left[\begin{array}{l}
v \\
w
\end{array}\right]\end{equation*}

The UAV assumes similar dynamics as in \cite{quadrotor}, where the states are center of mass position, linear velocity ($x$, $y$, and $z$ components), angle, and angular velocity (we consider pitch and roll but keep yaw fixed) or $X=[x,v_{x},\theta_{1}, \omega_{1}, y, v_{y}, \theta_{2}, \omega_{2}, z, v_{z}]$, and actions composed of thrust $U \in \mathbb{R}^{3\times1}$. The $A$ and $B$ matrices and their parameters are fully described in \cite{quadrotor}, where ${A}\in \mathbb{R}^{10\times10}$ and ${B}\in \mathbb{R}^{10\times3}$.

To solve the cost function \eqref{objfunction}, we use the mixed-integer nonlinear programming (MINLP) solver  $\lq$bonmin' \cite{bonami_algorithmic_2008}, as we need to consider both integer and continuous variables. Constraints and problem formulation were setup using CasADi \cite{Casadi} running on a laptop with an Intel Core i7‐8850H CPU, and NVIDIA Quadro P3200 GPU.

To collect data for training our RNN model for the UGV, we used the GMapping package \cite{Gmapping} to create a map of the environment, and then implemented the AMCL package \cite{AMCL} to track the robot's pose and receive the uncertainty covariances using this map with particle-filter SLAM (uncertainty outputs from the filter are considered as the $\lq$ground truth'). In the UAV case, we used the XIVO SLAM package \cite{xiao-semantic-mapping} to make localization and covariance estimations (XIVO uses an Extended-Kalman Filter). The Adamax optimizer was used for training and the Mean Squared Error (MSE) was implemented as the loss function for both RNN models (covariance matrices need to be converted to be positive semi-definite during usage, see \cite{Schperberg_rnn}). For training (for both models), we created 4 different maps with obstacles randomly distributed, where robots traverse about these maps via the SMPC. Note, that by definition of a particle-filter and an EKF, the former assumes non-Gaussian noise while the latter assumes Gaussian noise.

\subsection{Results}
\begin{figure}[!t]
    \centering
    \includegraphics[width=1.0\columnwidth]{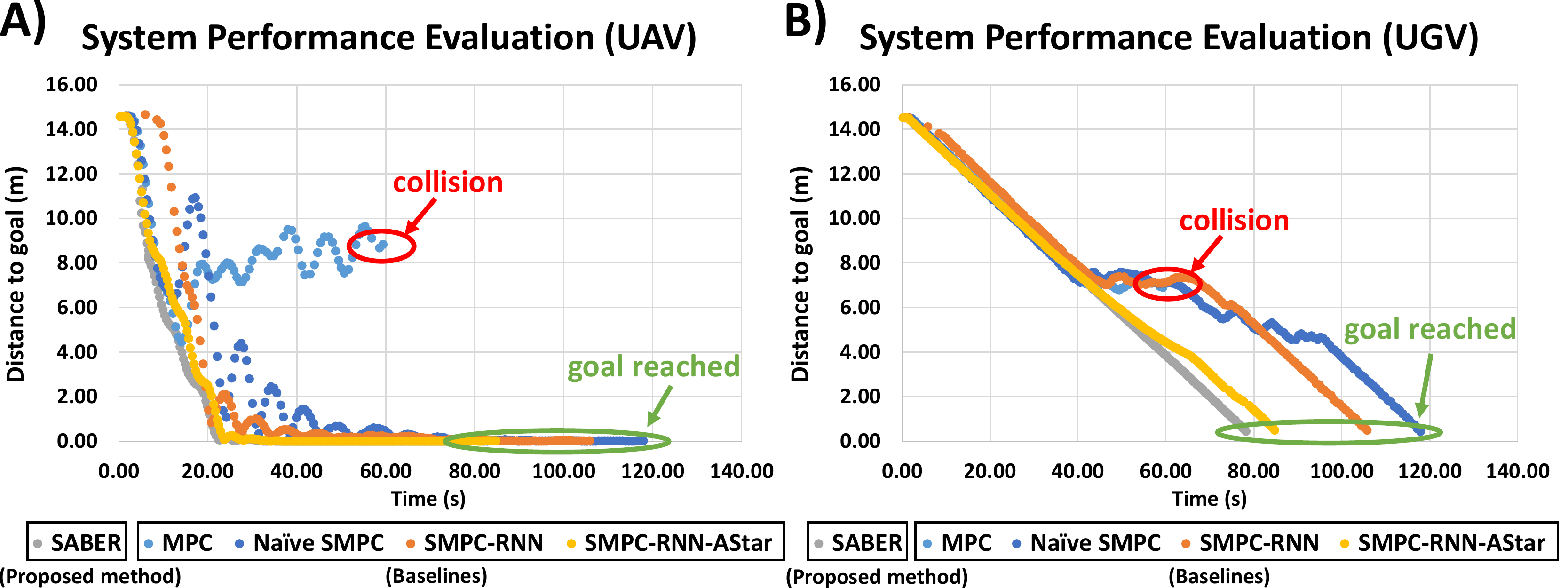}
    \caption{\textbf{System performance.} We compare the results of SABER (SMPC-RNN-DQN) using the same map as \textit{(D)} in Fig. \ref{fig_saber}) against baselines for the UAV \textit{(A)} and UGV \textit{(B)} with distance to goal vs time as our metric. See \ref{validation} for details.}
    \label{sys_perf}
\end{figure}

\subsubsection{Analysis of learning components and their significance}
The training loss for the RNN networks are shown in \Figref{rnn_loss}. The RNN networks for both SLAM algorithms were trained for 500 epochs (validation loss was observed to be close to the training loss), and shows a strong correlation between truth (output of SLAM in training) and predicted covariance. We also demonstrate that our RNN can model the behavior of covariance outputs generated by different sensor configurations (i.e., SLAM algorithms), as seen in (A) and (B) of \Figref{rnn_results} (e.g., we observed an increase in uncertainty for VIO SLAM when too close to obstacles, while seeing the opposite behavior for a particle-filter SLAM). Important to note, is that this result indicates that the RNN can model both non-Gaussian (particle-filter) and Gaussian (EKF) noise. By modeling the propagation of uncertainty of different measuring systems and integrating them into the SMPC prediction horizon through chance constraints, we ensure control values that avoid obstacle collision (without over avoidance) for each individual robot. 

In (B) of \Figref{fig_dqn_methods}, we show that during training of our DQN, the rewards would increase over a majority of the 35,000 episodes, reaching a steady-state at approximately 27,000 (except for the 5-robot team). In Table \ref{table1}, we also compared our DQN to other 2D baseline algorithms (we chose resolution settings for RRT, RRT-star, and A-star so that their path lengths were similar to the DQN, then evaluated their computation time). The results indicate that on average, the DQN had the lowest computation time and was comparable to A-star in terms of its path length to the goal (considered optimal for our map resolution). Unlike the baseline algorithms, our DQN also has the additional functionality of learning multi-agent semantic behavior--it was successful at moving the robots simultaneously to their goal points as shown in (C), and, as expected, stayed closer to each other when rewarding them based on close proximity as shown in (D) of \Figref{fig_dqn_methods}. Although it would be possible to modify the baseline planners to consider multi-agent planning, the learning-based approach considers potential semantics between agents that would be difficult to quantify and define beforehand within a cost function. 

\begin{table}[h!]
\caption{}
\label{table1}
\begin{tabular}{|P{1.13cm}|P{1.65cm}|P{1.22cm}|P{1.5cm}|P{1.14cm}|}
\hline
\multicolumn{5}{|c|}{\textit{High-level planner analysis (100 trials on map (D) of \Figref{fig_saber})}} \\
\hline
Planner& Path length(m)& std dev(m) & Solve time(s)& std dev(s) \\
\hline
RRT& 10.83& $\pm$1.28& 0.120& $\pm$0.091 \\
\hline
RRT-star& 10.40& $\pm$1.20& 0.084& $\pm$0.073 \\
\hline
A-star& 9.66& $\pm$0.00& 0.154& $\pm$0.002 \\
\hline
DQN& 9.68& $\pm$0.00& 0.051& $\pm$0.001 \\
\hline
\end{tabular}
\begin{tabular}{|P{1.3cm}|P{1.1cm}|P{0.9cm}|P{1.3cm}|P{0.8cm}|P{0.8cm}|}
\hline
\multicolumn{6}{|c|}{\textit{Average Computation time of SABER components (10 minutes of data)}} \\
\hline
Component& SMPC & RNN & VIO SLAM& PF SLAM & SABER \\
\hline
Solve time(s)& 0.0559& 0.0212 & 0.030& 0.073&0.193 \\
\hline
std dev(s)& $\pm$0.0187& $\pm$0.0077 & $\pm$0.003& $\pm$0.005 &$\pm$0.110  \\
\hline
\end{tabular}
\end{table}

The limitation of our DQN is that it's currently most capable in planning in 2D rather than 3D space, which helps lower training time and increase convergence (more complex DQN formulations may be necessary if the problem is either scaled to 3D, assumes a map bigger than 10$\times$10m, or uses more than 3 robots, see (B) of \Figref{fig_dqn_methods}). However, since we use an SMPC to avoid obstacles in 3D space using chance constraints, a 2D planner is sufficient for our application.

\subsubsection{Validation of the SABER algorithm}
\label{validation}
The complete planner is exemplified in \Figref{fig_saber} on a UGV-UAV team. We show that the SMPC of the UGV and UAV uses the state uncertainties estimated by an RNN to avoid colliding in obstacles in dense maps (see (A), and (B)). A special case is also shown in (C), where the SMPC of the UGV (without the DQN) reaches a local minima solution and is stuck behind the obstacle. However, when using the DQN's proposed path, the UGV can successfully reach its global goal. Note, that the SMPC and not the DQN considers both the dynamics of the robots and the uncertainty provided by their RNN models, thus, the actual path (shown in orange/purple) will differ slightly from the proposed DQN path (marked by triangles). We also evaluate the average computation time of the SABER algorithm (and its individual components) in Table \ref{table1}, showing a computation time of $\simeq{0.19}$ seconds/timestep. Lastly, in \Figref{sys_perf}, we verify that SABER performs best compared to several baselines (using distance to goal vs time as our metric on map (D) of \Figref{fig_saber}). The figure shows that MPC alone (no uncertainty is considered) causes an obstacle collision for the UGV and UAV (this likely occurred as the simulation contains noise, and the MPC is unaware of this noise during optimization). Both robots are able to get to their goals using a na\"ive stochastic MPC (uncertainty is considered by artificially inflating all obstacle boundaries), but due to over avoidance take longer to reach their goals. By adding our RNN to the SMPC allows both robots to reach their goals more quickly (uncertainty is now more accurately propagated within the SMPC prediction horizon). However, we observed that without a global planner, both robots run into local minima issues (i.e., robots would sometimes get stuck behind an obstacle for some time before reaching their goals). With a global planner (e.g., DQN, A-star), the robots reach their goals in the quickest manner (avoiding local minima issues). Although A-star and DQN provide near-optimal paths, the reason the DQN shows slightly better improvements (including better computation time) over A-star is because the DQN also accounts for multi-agent behavior, where robots were trained to stay close to each other when possible before reaching their respective goals (staying in close proximity decreases uncertainty via the Kalman filter).

\section{Conclusion}
In this work, we demonstrated that the SABER algorithm (which combines several fields of robotics including controls, vision, and machine learning into a single framework) is computationally feasible ($\simeq{0.19}$ seconds/timestep) and plans paths for heterogeneous robots to reach a global goal while satisfying diverse dynamics, constraints, and consideration of uncertainty. In future work, we plan to relax the assumption of a perfect object detection system and will focus on expanding our DQN to consider more complex behavior and tasks.

\bibliographystyle{IEEEtran}
\bibliography{bibliography}

\end{document}